\documentclass{article}
\usepackage{spconf,amsmath,graphicx,amssymb,amsthm,comment,bm,cite,url}
\usepackage{array}
\allowdisplaybreaks

\newcommand{\refeq}[1]{(\ref{eq:#1})}
\newcommand{\refeqs}[2]{(\ref{eq:#1}) and (\ref{eq:#2})}

\newcommand{\refsec}[1]{Section \ref{sec:#1}}
\newcommand{\refsubsec}[1]{\ref{subsec:#1}}

\newcommand{\reffig}[1]{Fig. \ref{fig:#1}}

\def\Vec#1{\boldsymbol{\mathbf{#1}}}

\def\x{\Vec{x}}

\def\sh{\boldsymbol{\mathsf h}}
\def\sW{\boldsymbol{\mathsf W}}
\def\sV{\boldsymbol{\mathsf V}}
\def\sb{\boldsymbol{\mathsf b}}
\def\sd{\boldsymbol{\mathsf d}}

\def\encdis{q_{\phi}}
\def\decdis{p_{\theta}}
\def\auxdis{r_{\psi}}

\def\CS{{\mathcal S}}

\def\z{\Vec{z}}

\def\0{{\mathbf 0}}
\def\I{{\mathbf I}}
\def\vmu{{\boldsymbol \mu}}
\def\vsigma{{\boldsymbol \sigma}}
\def\vepsilon{{\boldsymbol \epsilon}}

\makeatletter
\newcommand{\thickhline}{%
    \noalign {\ifnum 0=`}\fi \hrule height 1pt
    \futurelet \reserved@a \@xhline
}
\newcolumntype{"}{@{\hskip\tabcolsep\vrule width 1pt\hskip\tabcolsep}}
\makeatother

\title{
ACVAE-VC: 
Non-parallel many-to-many voice conversion with \\
auxiliary classifier variational autoencoder
}
\name{
Hirokazu Kameoka, Takuhiro Kaneko, Kou Tanaka, Nobukatsu Hojo
}
\address{NTT Communication Science Laboratories, NTT Corporation, Japan}
\begin{document}
%
\maketitle
\begin{abstract}
This paper proposes a non-parallel many-to-many voice conversion (VC) method using a variant of the conditional variational autoencoder (VAE) called an auxiliary classifier VAE (ACVAE). The proposed method has three key features. First, it adopts fully convolutional architectures to construct the encoder and decoder networks so that the networks can learn conversion rules that capture time dependencies in the acoustic feature sequences of source and target speech. Second, it uses an information-theoretic regularization for the model training to ensure that the information in the attribute class label will not be lost in the conversion process. With regular CVAEs, the encoder and decoder are free to ignore the attribute class label input. This can be problematic since in such a situation, the attribute class label will have little effect on controlling the voice characteristics of input speech at test time. Such situations can be avoided by introducing an auxiliary classifier and training the encoder and decoder so that the attribute classes of the decoder outputs are correctly predicted by the classifier. Third, it avoids producing buzzy-sounding speech at test time by simply transplanting the spectral details of the input speech into its converted version. Subjective evaluation experiments revealed that this simple method worked reasonably well in a non-parallel many-to-many speaker identity conversion task.
\end{abstract}
\section{Introduction}
\label{sec:intro}

Voice conversion (VC) is a technique for converting 
para/non-linguistic information
contained in a given utterance without changing the linguistic information.
This technique can be applied to various
tasks such as speaker-identity modification for text-to-speech
(TTS) systems \cite{Kain1998}, speaking assistance \cite{Kain2007,Nakamura2012}, speech enhancement \cite{Inanoglu2009,Turk2010,Toda2012}, and pronunciation conversion \cite{Kaneko2017c}.

One widely studied VC framework involves 
Gaussian mixture model (GMM)-based approaches \cite{Stylianou1998,Toda2007,Helander2010}.  
Recently, neural network (NN)-based frameworks based on restricted Boltzmann machines \cite{Chen2014,Nakashika2014a}, feed-forward deep NNs \cite{Desai2010,Mohammadi2014}, recurrent NNs \cite{Nakashika2014b,Sun2015}, variational autoencoders (VAEs) \cite{Blaauw2016,Hsu2016,Hsu2017} and generative adversarial nets (GANs) \cite{Kaneko2017c}, and an exemplar-based framework based on non-negative matrix factorization (NMF) \cite{Takashima2013,Wu2014} have also attracted particular attention. 
While many VC methods including those mentioned above require accurately aligned parallel data of
source and target speech, in general scenarios, 
collecting parallel utterances can be a costly and time-consuming process. 
Even if we were able to collect parallel utterances, we typically need to perform 
automatic time alignment procedures, which becomes relatively
difficult when there is a large acoustic gap between the
source and target speech. Since many frameworks are weak with respect to
the misalignment found with parallel data, careful pre-screening and
manual correction is often required to make these frameworks work reliably.
To sidestep these issues, this paper aims to develop a non-parallel 
VC method that requires no parallel utterances, transcriptions, or time alignment procedures.

The quality and conversion effect obtained with non-parallel methods
are generally poorer than with methods using parallel data 
since there is a disadvantage related to the training condition. 
Thus, it would be challenging to achieve 
as high a quality and conversion effect with non-parallel methods
as with parallel methods. 
Several non-parallel methods have already been proposed \cite{Hsu2016,Hsu2017,Xie2016,Kinnunen2017}.
For example, a method using automatic speech recognition (ASR) was proposed in \cite{Xie2016}
where the idea is to convert input speech under a restriction, namely that the posterior state probability of the acoustic model of an ASR system is preserved. 
Since the performance of this method depends heavily on the quality of the acoustic model of ASR,
it can fail to work if ASR does not function reliably. 
A method using i-vectors \cite{Dehak2011}, which is known to be a powerful feature for speaker verification, was proposed in \cite{Kinnunen2017} where the idea is to
shift the acoustic features of input speech towards target speech in the i-vector space so that
the converted speech is likely to be recognized as the target speaker by a speaker recognizer.
While this method is also free of parallel data, one limitation is that 
it is applicable only to speaker identity conversion tasks.

Recently, a framework based on conditional variational autoencoders (CVAEs) \cite{Kingma2014a,Kingma2014b} was proposed in \cite{Hsu2016,YSaito2018bshort}.
As the name implies, VAEs are a probabilistic counterpart of autoencoders (AEs), consisting of encoder and decoder networks.
Conditional VAEs (CVAEs) \cite{Kingma2014b} are an extended version of VAEs 
with the only difference being that 
the encoder and decoder networks take an attribute class label $c$ as an additional input. 
By using acoustic features associated with attribute labels 
as the training examples, 
the networks learn how to convert an attribute of source speech to a target attribute according to the attribute label fed into the decoder. 
While this VAE-based VC approach is notable in that it is completely free of parallel data and works even with unaligned corpora, there are three major drawbacks.
Firstly, 
the devised networks are designed to 
produce acoustic features frame-by-frame, which makes it difficult to
learn time dependencies in the acoustic feature sequences of source and target speech. 
Secondly, 
one well-known problem as regards VAEs is that outputs from the decoder tend to be oversmoothed. 
This can be problematic for VC applications since it usually results in poor quality buzzy-sounding speech. 
One natural way of alleviating the oversmoothing effect in VAEs would be to 
use the VAE-GAN framework \cite{Larsen2015short}. 
A non-parallel VC method based on this framework has already been proposed in \cite{Hsu2017}.
With this method, 
an adversarial loss derived using a GAN discriminator is incorporated into the training loss to 
make the decoder outputs of a CVAE indistinguishable from real speech features.
While this method is able to produce more realistic-sounding speech than the regular 
VAE-based method \cite{Hsu2016}, 
as will be shown in \refsec{experiments}, the audio quality 
and conversion effect are still limited. 
Thirdly,  
in the regular CVAEs, the encoder and decoder 
are free to ignore the additional input $c$ by finding networks that 
can reconstruct any data without using $c$. 
In such a situation, the attribute class label $c$
will have little effect on controlling 
the voice characteristics of the input speech.

To overcome these drawbacks and limitations, in this paper we describe three modifications to 
the conventional VAE-based approach. 
First, we adopt fully convolutional architectures to design the encoder and decoder networks
so that the networks can learn conversion rules that capture short- and long-term dependencies
in the acoustic feature sequences of source and target speech.
Secondly, we propose simply transplanting the spectral details of input speech 
into its converted version at test time to avoid producing buzzy-sounding speech. 
We will show in \refsec{experiments} that this simple method 
works considerably better than the VAE-GAN framework \cite{Hsu2017} 
in terms of audio quality.
Thirdly, we propose using an information-theoretic regularization
for the model training to ensure that 
the attribute class information will not be lost in the conversion process. 
This can be done by introducing an auxiliary classifier whose role is to predict to which 
attribute class an input acoustic feature sequence belongs and 
by training the encoder and decoder 
so that the attribute classes of the decoder outputs are correctly predicted by the classifier.
We call the present VAE variant an auxiliary classifier VAE (or ACVAE). 

\section{VAE voice conversion}

\subsection{Variational Autoencoder (VAE)}
\label{subsec:vae}

VAEs \cite{Kingma2014a,Kingma2014b} are stochastic neural network models consisting of encoder and decoder networks.
The encoder network generates a set of parameters for the conditional distribution $\encdis(\z|\x)$ of a latent space variable $\z$ given input data $\x$, whereas 
the decoder network generates a set of parameters for the conditional distribution $\decdis(\x|\z)$ of the data $\x$ given the latent space variable $\z$. 
Given a training dataset $\CS = \{ \x_m \}_{m=1}^{M}$, 
VAEs learn the parameters of 
the entire network so that the encoder distribution $\encdis(\z|\x)$ becomes consistent with the posterior $\decdis(\z|\x)\propto \decdis(\x|\z)p(\z)$. 
By using Jensen's inequality, the log marginal distribution of data $\x$ can be lower-bounded by
\begin{align}
\log \decdis(\x) &
= \log \int \encdis(\z|\x) \frac{\decdis(\x|\z)p(\z)}{\encdis(\z|\x)} d\z 
\nonumber\\
&\ge
\int \encdis(\z|\x) \log \frac{\decdis(\x|\z)p(\z)}{\encdis(\z|\x)} d\z
\label{eq:lowerbound}\\
&=
\mathbb{E}_{\z\sim \encdis(\z|\x)}[\log \decdis(\x|\z)] 
- {\rm KL}[ \encdis(\z|\x) \| p(\z) ],
\nonumber
\end{align}
where the difference between the left- and right-hand sides of this inequality 
is equal to the Kullback-Leibler divergence ${\rm KL}[\encdis(\z|\x) \| \decdis(\z|\x)]$,
which is minimized when
\begin{align}
\encdis(\z|\x) = \decdis(\z|\x).
\end{align}
This means we can make $\encdis(\z|\x)$ and $\decdis(\z|\x)\propto \decdis(\x|\z)p(\z)$ 
consistent by maximizing the lower bound of \refeq{lowerbound}.
One typical way of modeling $\encdis(\z|\x)$, $\decdis(\x|\z)$ and $p(\z)$ is to assume
Gaussian distributions
\begin{align}
\encdis(\z|\x) &= \mathcal{N}(\z|\vmu_{\phi}(\x), {\rm diag}(\vsigma_{\phi}^2(\x))),
\label{eq:q(z|s)}
\\
\decdis(\x|\z) &= \mathcal{N}(\x|\vmu_{\theta}(\z), {\rm diag}(\vsigma_{\theta}^2(\z))),
\label{eq:p(s|z)}
\\
p(\z) &= \mathcal{N}(\z|{\bf 0},{\bf I}),
\label{eq:p(z)}
\end{align}
where $\vmu_{\phi}(\x)$ and $\vsigma_{\phi}^2(\x)$ are the outputs of an encoder network with parameter $\phi$,
and $\vmu_{\theta}(\z)$ and $\vsigma_{\theta}^2(\z)$ are the outputs of a decoder network with parameter $\theta$.
The first term of the lower bound can be interpreted as an autoencoder reconstruction error.
By using a reparameterization $\z = \vmu_{\phi}(\x) + \vsigma_{\phi}(\x) \odot \vepsilon$ with 
$\vepsilon \sim \mathcal{N}(\vepsilon|\0,\I)$, 
sampling $\z$ from $\encdis(\z|\x)$ can be replaced by
sampling $\vepsilon$ from the distribution, which is independent of $\theta$.
This allows us to compute the gradient of the lower bound with respect to $\theta$
by using a Monte Carlo approximation of the expectation $\mathbb{E}_{\z\sim\encdis(\z|\x)}[\cdot]$. 
The second term is given as the negative KL divergence 
between $\encdis(\z|\x)$ and $p(\z)=\mathcal{N}(\z|\0,\I)$. This term can be interpreted as a regularization term that 
forces each element of the encoder output to be uncorrelated and normally distributed.

Conditional VAEs (CVAEs) \cite{Kingma2014b} are an extended version of VAEs 
with the only difference being that
the encoder and decoder networks can take an auxiliary variable 
$c$ as an additional input. With CVAEs, \refeqs{q(z|s)}{p(s|z)} are replaced with
\begin{align}
\encdis(\z|\x,c) &= \mathcal{N}(\z|\vmu_{\phi}(\x,c), {\rm diag}(\vsigma_{\phi}^2(\x,c))),
\label{eq:q(z|s,c)}
\\
\decdis(\x|\z,c) &=
\mathcal{N}(\x|\vmu_{\theta}(\z,c), {\rm diag}(\vsigma_{\theta}^2(\z,c))),
\label{eq:p(s|z,c)}
\end{align}
and the variational lower bound to be maximized becomes
\begin{align}
\mathcal{J}(\phi,\theta) 
=
&
\mathbb{E}_{
(\x,c)\sim p_{\rm D}(\x,c)
}\big[ 
\mathbb{E}_{\z\sim q(\z|\x,c)}[\log p(\x|\z,c)] 
\nonumber\\
&~~~~~~~~~~~~~~~~~~~~~~~~~~~~~~
- {\rm KL}[ q(\z|\x,c) \| p(\z) ]
\big],
\end{align}
where $\mathbb{E}_{(\x,c)\sim p_{\rm D}(\x,c)}[\cdot]$ denotes the sample mean over the training examples $\{\x_m,c_m\}_{m=1}^{M}$.

\begin{figure*}[t!]
\centering
  \begin{minipage}{.75\linewidth}
  \centerline{\includegraphics[width=.98\linewidth]{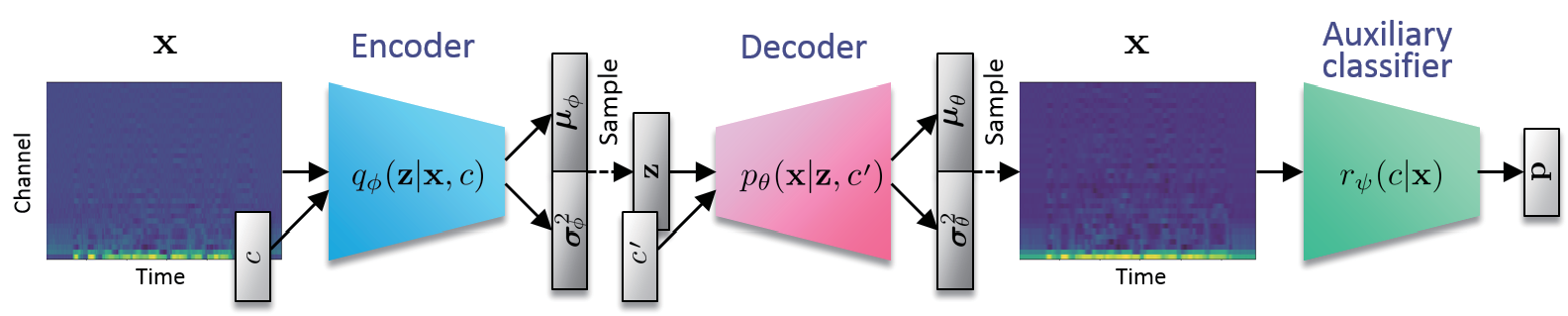}}
  \vspace{-2ex}
  \caption{Illustration of ACVAE-VC.}
  \label{fig:acvae-vc}
  \end{minipage}
\end{figure*}

\subsection{Non-parallel voice conversion using CVAE}

By letting $\x\in\mathbb{R}^{Q}$ and $c$ be an acoustic feature vector and an attribute class label, 
a non-parallel VC problem can be formulated using the CVAE \cite{Hsu2016,Hsu2017}. 
Given a training set of acoustic features with attribute class labels $\{\x_m, c_m\}_{m=1}^M$, 
the encoder learns to map an input acoustic feature $\x$ and an attribute class label $c$
to a latent space variable $\z$ (expected to represent phonetic information) 
and then the decoder reconstructs an acoustic feature $\hat{\x}$ 
conditioned on the encoded latent space variable
$\z$ and the attribute class label $c$. 
At test time, we can generate a converted feature by feeding an acoustic feature of the input speech
into the encoder and a target attribute class label into the decoder.   

\section{Proposed method}
\label{sec:acvae-vc}
\subsection{Fully Convolutional VAE}
\label{subsec:fcn}

While the model in \cite{Hsu2016,Hsu2017} is designed to convert acoustic features frame-by-frame
and fails to learn conversion rules that reflect time-dependencies in acoustic feature sequences, 
we propose extending it to a sequential version to overcome this limitation.
Namely, we devise a CVAE that 
takes an acoustic feature sequence instead of a single-frame acoustic feature as an input and 
outputs an acoustic feature sequence of the same length. 
Hence, in the following we assume that $\x \in\mathbb{R}^{Q\times N}$ is 
an acoustic feature sequence of length $N$. 
While RNN-based architectures are a natural choice for modeling time series data, 
we use fully convolutional networks to design $\encdis$ and $\decdis$, 
as detailed in \refsubsec{netarch}. 

\subsection{Auxiliary Classifier VAE} 
\label{subsec:acvae}

We hereafter assume that a class label comprises
one or more categories, each consisting of multiple classes.
We thus represent $c$ as a concatenation of one-hot vectors,
each of which is filled with 1 at the index of a class in a certain
category and with 0 everywhere else. For example, if
we consider speaker identities as the only class category,
$c$ will be represented as a single one-hot vector, where each
element is associated with a different speaker.

The regular CVAEs impose no restrictions on the manner in which the
encoder and decoder may use the attribute class label $c$.
Hence, the encoder and decoder are free to ignore $c$ by finding 
distributions satisfying $\encdis(\z|\x,c)=\encdis(\z|\x)$ and $\decdis(\x|\z,c)=\decdis(\x|\z)$.
This can occur for instance when the encoder and decoder have 
sufficient capacity to reconstruct any data without using $c$.
In such a situation, $c$ will have little effect on 
controlling the voice characteristics of input speech.
To avoid such situations, we introduce an information-theoretic regularization 
\cite{Chen2016} to assist the decoder output to be correlated 
as far as possible with $c$.

The mutual information for $\x\sim \decdis(\x|\z,c)$ and $c$ 
conditioned on $\z$ 
can be written as
\begin{align}
&I(c,\x|\z) \nonumber\\
&= \mathbb{E}_{c\sim p(c), \x\sim \decdis(\x|\z,c),c'\sim p(c|\x)}[\log p(c'|\x)] + H(c),
\label{eq:MI}
\end{align}
where $H(c)$ represents the entropy of $c$, which can be considered a constant term.
In practice, $I(c,\x|\z)$ is hard to optimize directly since it requires
access to the posterior $p(c|\x)$. Fortunately, we can obtain a lower bound of 
the first term of $I(c;\x|\z)$ 
by introducing an auxiliary distribution $r(c|\x)$
\begin{align}
&\mathbb{E}_{c\sim p(c), \x\sim \decdis(\x|\z,c),c'\sim p(c|\x)}[\log p(c'|\x)]
\nonumber\\
=&\mathbb{E}_{c\sim p(c), \x\sim \decdis(\x|\z,c),c'\sim p(c|\x)}
\left[\log \frac{r(c'|\x)p(c'|\x)}{r(c'|\x)}\right] 
\nonumber\\
\ge& \mathbb{E}_{c\sim p(c), \x\sim \decdis(\x|\z,c),c'\sim p(c|\x)}[\log r(c'|\x)] 
\nonumber\\
=&
\mathbb{E}_{c\sim p(c), \x\sim \decdis(\x|\z,c)}[\log r(c|\x)].
\label{eq:MI-LB}
\end{align}
This technique of lower bounding mutual information is known as 
variational information maximization \cite{Barber2003}. 
The last line of \refeq{MI-LB} follows from the lemma presented in \cite{Chen2016}.
The equality holds in \refeq{MI-LB} when $r(c|\x)=p(c|\x)$.
Hence, maximizing the lower bound \refeq{MI-LB} with respect to $r(c|\x)$ corresponds to 
approximating $p(c|\x)$ by $r(c|\x)$ as well as approximating $I(c,\x|\z)$ by this lower bound.
We can therefore indirectly increase $I(c,\x|\z)$ by increasing the lower bound with respect to
$\decdis(\x|\z,c)$ and $r(c|\x)$. One way to do this involves 
expressing $r(c|\x)$ using an NN and training it along with $\encdis(\z|\x,c)$ and $\decdis(\x|\z,c)$. 
Hereafter, we use $r_{\psi}(c|\x)$ to denote the auxiliary classifier NN with parameter $\psi$.
As detailed in \refsubsec{netarch},
we also design the auxiliary classifier using a fully convolutional network, which
takes an acoustic feature sequence as the input
and generates a sequence of class probabilities.
The regularization term that we would like to maximize with respect to $\phi$, $\theta$ and $\psi$
becomes
\begin{align}
&\mathcal{L}(\phi,\theta,\psi) 
\\
&= 
\mathbb{E}_{(\tilde{c},\tilde{x})\sim p_D(\tilde{\x},\tilde{c}),\encdis(\z|\tilde{\x},\tilde{c})}
\big[
\mathbb{E}_{c\sim p(c), \x\sim \decdis(\x|\z,c)}[\log r_{\psi}(c|\x)]
\big],
\nonumber
\end{align}
where $\mathbb{E}_{(\tilde{\x},\tilde{c})\sim p_D(\tilde{\x},\tilde{c})}[\cdot]$ 
denotes the sample mean over the training examples $\{\tilde{\x}_m,\tilde{c}_m\}_{m=1}^{M}$.
Fortunately, we can use the same reparameterization trick as in 
\refsubsec{vae} to compute the gradients of $\mathcal{L}(\phi,\theta,\psi)$
with respect to $\phi$, $\theta$ and $\psi$.
Since we can also use 
the training examples $\{\tilde{\x}_m,\tilde{c}_m\}_{m=1}^{M}$ to train 
the auxiliary classifier $\auxdis(c|\x)$, we include 
the cross-entropy 
\begin{align}
\mathcal{I}(\psi) = \mathbb{E}_{(\tilde{\x},\tilde{c})\sim p_D(\tilde{\x},\tilde{c})}
[
\log \auxdis(\tilde{c}|\tilde{\x})
],
\end{align}
in our training criterion. 
The entire training criterion is thus given by
\begin{align}
\mathcal{J}(\phi,\theta) 
+ \lambda_{\mathcal{L}}\mathcal{L}(\phi,\theta,\psi) 
+ \lambda_{\mathcal{I}}\mathcal{I}(\psi),
\end{align}
where $\lambda_{\mathcal{L}}\ge 0$ and $\lambda_{\mathcal{I}}\ge 0$ are 
regularization parameters, which 
weigh the importances of the regularization terms 
relative to the VAE training criterion $\mathcal{J}(\phi,\theta)$.

While the idea of using the auxiliary classifier for 
GAN-based image synthesis \cite{Odena2017,Choi2017short} and voice conversion \cite{Kameoka2018} 
has already been proposed, to the best of our knowledge, it has yet to be proposed 
for use with the VAE framework.
We call the present VAE variant an auxiliary classifier VAE (or ACVAE).

\subsection{Conversion Process}
\label{subsec:conversion}

Although it would be interesting to develop an end-to-end model 
by directly using a time-domain signal or a magnitude spectrogram 
as $\x$, in this paper we use a sequence of 
mel-cepstral coefficients \cite{Fukada1992} 
computed from a spectral envelope sequence obtained using WORLD \cite{Morise2016}. 

After training $\phi$ and $\theta$, we can convert $\x$ with
\begin{align}
\hat{\x} = \vmu_{\theta}(\vmu_{\phi}(\x, c), \hat{c}),
\end{align}
where $c$ and $\hat{c}$ denote the source and target attribute class labels, respectively.
A na\"ive way of obtaining a time-domain signal is to simply use $\hat{\x}$ to 
reconstruct a signal with a vocoder.
However, the converted feature sequence $\hat{\x}$ obtained with this procedure
tended to be over-smoothed as with other conventional VC methods, 
resulting in buzzy-sounding synthetic speech. 
This was also the case with the reconstructed feature sequence 
\begin{align}
\bar{\x} = \vmu_{\theta}(\vmu_{\phi}(\x, c), {c}).
\end{align}
This oversmoothing effect was caused by the Gaussian assumptions on the encoder and decoder distributions: 
Under the Gaussian assumptions, 
the encoder and decoder networks learn to fit the decoder outputs to the inputs in an expectation sense. 
Instead of directly using $\hat{\x}$ to reconstruct a signal,
a reasonable way of avoiding this over-smoothing effect is to transplant the spectral details of the input speech into its converted version. 
By using $\hat{\x}$ and $\bar{\x}$, 
we can obtain a sequence of spectral gain functions by dividing 
$F(\hat{\x})$ by $F(\bar{\x})$ where $F$ denotes a transformation 
from an acoustic feature sequence to a spectral envelope sequence. 
Once we obtain the spectral gain functions, 
we can reconstruct a time-domain signal 
by multiplying 
the spectral envelope of the input speech by
the spectral gain function frame-by-frame
and resynthesizing the signal using a WORLD vocoder.
Alternatively, we can adopt the vocoder-free 
direct waveform modification method \cite{Kobayashi2016}, which 
consists of transforming the spectral gain functions into time-domain impulse responses
and convolving the input signal with the obtained filters.

\subsection{Network Architectures}
\label{subsec:netarch}

\begin{figure*}[t!]
\centering
  \centerline{\includegraphics[width=.65\linewidth]{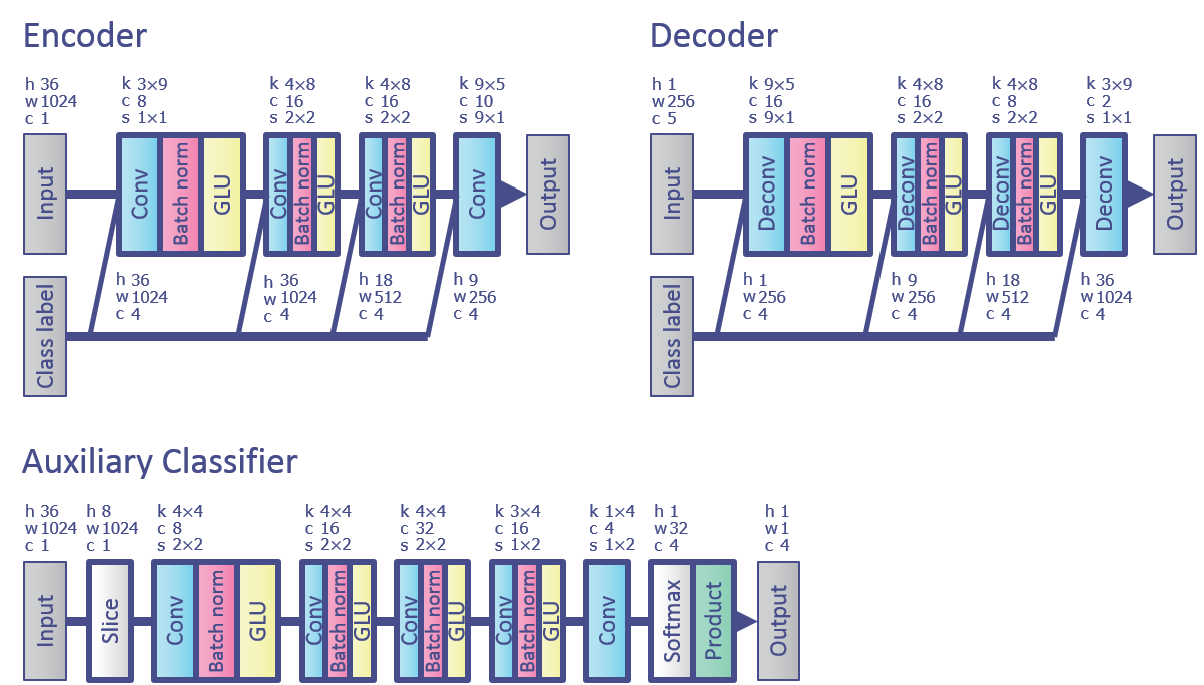}}
  \vspace{-0ex}
  \caption{Network architectures of the encoder, decoder and auxiliary classifier. Here, the input and output of each of the networks are interpreted as images, where ``h'', ``w'' and ``c'' denote the height, width and channel number, respectively. ``Conv'', ``Batch norm'', ``GLU'', ``Deconv'' ``Softmax'' and ``Product'' denote convolution, batch normalization, gated linear unit, transposed convolution, softmax, and product pooling layers, respectively. ``k'', ``c'' and ``s'' denote the kernel size, output channel number and stride size of a convolution layer, respectively. Note that all the networks are fully convolutional with no fully connected layers, thus allowing inputs to have arbitrary sizes.}
\label{fig:netarch}
\end{figure*}

{\noindent \bf Encoder/Decoder:}
We use 2D CNNs to design the encoder and the decoder networks and the auxiliary classifier network 
by treating ${\x}$ as an image of size $Q \times N$ with $1$ channel.
Specifically, we use a gated CNN \cite{Dauphin2017}, which 
was originally introduced to model word sequences for language modeling 
and was shown to outperform long short-term memory (LSTM) language models trained in a similar setting. 
We previously employed gated CNN architectures for voice conversion \cite{Kaneko2017c,Kaneko2017d,Kameoka2018} and monaural audio source separation \cite{Li2018}, and their effectiveness has already been confirmed. 
In the encoder, the output of the
$l$-th hidden layer, $\sh_l$,   
is described as a linear projection 
modulated by an output gate
\begin{align}
\sh_{l-1}' &= [\sh_{l-1};\Vec{c}_{l-1}],
\label{eq:dec_glu_1}
\\
\sh_l &= (\sW_l * \sh_{l-1}' + \sb_l) \odot 
\sigma
(\sV_l * \sh_{l-1}' + \sd_l),
\label{eq:dec_glu_2}
\end{align}
where 
$\sW_l \in \mathbb{R}^{D_l\!\times\! D_{l-1}\!\times\! Q_l\!\times\! N_{l}}$, 
$\sb_l\in \mathbb{R}^{D_l}$, 
$\sV_l\in \mathbb{R}^{D_l\! \times\! D_{l-1}\!\times\! Q_l \!\times\! N_{l}}$ and 
$\sd_l\in \mathbb{R}^{D_l}$ are
the encoder network parameters $\phi$, and 
$\sigma$ denotes the elementwise sigmoid function.
Similar to LSTMs,
the output gate multiplies each element of 
$\sW_l * \sh_{l-1} + \sb_l$
and control what information should be propagated through the hierarchy of layers.
This gating mechanism is called a gated linear unit (GLU).
Here, $[\sh_{l};\Vec{c}_{l}]$ means the concatenation of 
$\sh_{l}$ and $\Vec{c}_{l}$
along the channel dimension, and
$\Vec{c}_l$ is 
a 3D array consisting of a $Q_l$-by-$N_l$ tiling of copies of $c$ in the time dimensions.
The input into the 1st layer of the encoder is $\sh_0 = {\x}$.
The outputs of the final layer are 
given as regular linear projections
\begin{align}
\vmu_{\phi} &= \sW_L * \sh_{L-1}' + \sb_L,\\
\log \vsigma_{\phi}^2 &= \sV_L * \sh_{L-1}' + \sd_L.
\end{align}
The decoder network is constructed as described below:
\begin{align}
\sh_0 &= \z,\nonumber\\
\sh'_{l-1} &= [\sh_{l-1};\Vec{c}_{l-1}],
\nonumber\\
\sh_l &= (\sW'_l * \sh'_{l-1} + \sb'_l) \odot 
\sigma
(\sV'_l * \sh'_{l-1} + \sd'_l),\nonumber\\
\vmu_{\theta} &= \sW'_L * \sh'_{L-1} + \sb'_L,\nonumber\\
\log \vsigma_{\theta}^2 &= \sV'_L * \sh'_{L-1} + \sd'_L,\nonumber
\end{align}
where $\sW'_l \in \mathbb{R}^{D_l\!\times\! D_{l-1}\!\times\!Q_l\times\! N_l}$, 
$\sb'_l\in \mathbb{R}^{D_l}$, 
$\sV'_l\in \mathbb{R}^{D_l\! \times\! D_{l-1}\! \times\! Q_l \! \times N_l}$ 
and $\sd'_l\in \mathbb{R}^{D_l}$ are the decoder network parameters $\theta$.
See \refsec{experiments} for more details.
It should be noted that since
the entire architecture is fully convolutional with no fully-connected layers,
it can take an entire sequence with an arbitrary length as an input and
generate an acoustic feature sequence of the same length.

{\noindent \bf Auxiliary Classifier:}
We also design an auxiliary classifier 
using a gated CNN, which takes an acoustic feature sequence
$\x$ and produces a sequence of class probability distributions
that shows how likely each segment of $\x$ is to belong to
attribute $c$. The output of the $l$-th layer of the classifier 
is given as 
\begin{align}
\sh_l &= (\sW''_l * \sh_{l-1} + \sb''_l) \odot 
\sigma
(\sV''_l * \sh_{l-1} + \sd''_l),
\end{align}
where $\sW''_l \in \mathbb{R}^{D_l\!\times\! D_{l-1}\!\times\!Q_l\times\! N_l}$, 
$\sb''_l\in \mathbb{R}^{D_l}$, 
$\sV''_l\in \mathbb{R}^{D_l\! \times\! D_{l-1}\! \times\! Q_l \! \times N_l}$ 
and $\sd''_l\in \mathbb{R}^{D_l}$ are the auxiliary classifier network parameters $\psi$.
The final output $r_{\psi}(c|\x)$ is given by
the product of all the elements of $\sh_L$. 
See \refsec{experiments} for more details.

\section{Experiments}
\label{sec:experiments}

\begin{figure}[t!]
\centering
\begin{minipage}{.75\linewidth}
  \centerline{\includegraphics[height=6.2cm]{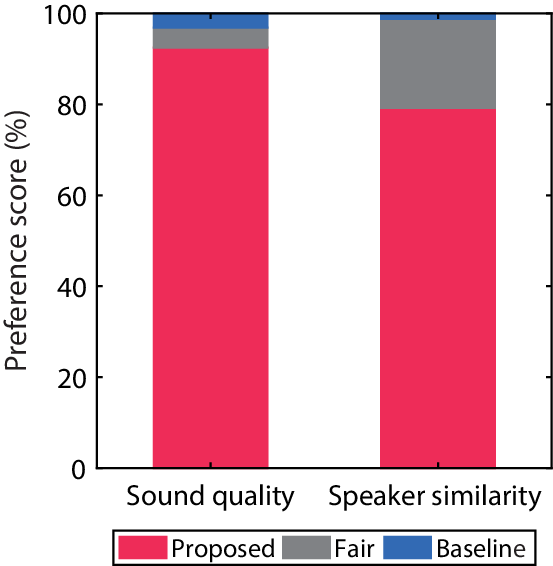}}
\caption{Results of the AB test for sound quality and the ABX test for speaker similarity.}
\label{fig:abx}
\end{minipage}
\end{figure}

To confirm the performance of our proposed method, 
we conducted subjective evaluation experiments involving 
a non-parallel many-to-many speaker identity conversion task. 
We used the Voice Conversion Challenge (VCC) 2018 dataset \cite{Lorenzo-Trueba2018short},
which consists of recordings of six female and six male US English speakers. 
We used a subset of speakers for training and evaluation. 
Specifically, we selected two female speakers, `VCC2SF1' and `VCC2SF2', 
and two male speakers, `VCC2SM1' and `VCC2SM2'. 
Thus, $c$ is represented as a four-dimensional one-hot vector and in total
there were twelve different combinations of source and target speakers.
The audio files for each speaker were manually
segmented into 116 short sentences (each about 7 minutes long)
where 81 and 35 sentences (each, respectively, about 5 and 2 minutes long) were provided
as training and evaluation sets, respectively. 
All the speech signals were sampled at 22050 Hz. 
For each utterance, a spectral envelope,
a logarithmic fundamental frequency (log $F_0$), and
aperiodicities (APs) were extracted every 5 ms using the
WORLD analyzer \cite{Morise2016}. 36 mel-cepstral coefficients (MCCs) were then extracted from 
each spectral envelope. 
The $F_0$ contours were converted using the logarithm Gaussian normalized
transformation described in \cite{Liu2007}. The aperiodicities were used directly without
modification. The network configuration is shown in detail in \reffig{netarch}.
The signals of the converted speech were obtained using the method described in \refsubsec{conversion}.

We chose the VAEGAN-based approach \cite{Hsu2017} for comparison with our experiments. 
Although we would have liked to replicate the implementation of this method exactly, 
we made our own design choices because certain details of the network configuration and hyperparameters were missing.
We conducted an AB test to compare the sound quality of the converted speech samples
and an ABX test to compare the similarity to the target speaker of the converted speech samples, where ``A'' and ``B'' were converted speech samples obtained with the proposed and baseline methods and ``X'' was a real speech sample obtained from a target speaker.
With these listening tests, 
``A'' and ``B'' were presented in random orders to eliminate bias in the order of stimuli. 
Eight listeners participated in our listening tests. 
For the AB test of sound quality, 
each listener was presented \{``A'',``B''\} $\times$ 20 utterances, 
and for the ABX test of speaker similarity,
each listener was presented \{``A'',``B'',``X''\} $\times$ 24 utterances.
Each listener was then asked to select ``A'', ``B'' or ``fair'' for each utterance.
The results are shown in \reffig{abx}.
As the results reveal, 
the proposed method significantly outperformed the baseline method in terms of 
both sound quality and speaker similarity.
Audio samples are provided at 
http://www.kecl.ntt.co.jp/people/kameoka.hirokazu/Demos/ acvae-vc/.

\section{Conclusions}

This paper proposed a non-parallel many-to-many VC method  
using a VAE variant called an auxiliary classifier VAE (ACVAE). 
The proposed method has three key features.
First, we adopted fully convolutional architectures
to construct the encoder and decoder networks so that the networks could learn conversion rules that capture 
time dependencies in the acoustic feature sequences of source and target speech.
Second, we proposed using an information-theoretic regularization for the model training 
to ensure that the information in the latent attribute label would not be lost in the generation process.
With regular CVAEs, the encoder and decoder are free to ignore the attribute class label input. 
This can be problematic since in such a situation, the attribute class label input
will have little effect on controlling the voice characteristics of the input speech. 
To avoid such situations, we proposed introducing an auxiliary classifier and 
training the encoder and decoder so that the attribute classes of the decoder outputs are correctly predicted by the classifier.
Third, to avoid producing buzzy-sounding speech at test time, 
we proposed simply transplanting the spectral details of the input speech into its converted version. 
Subjective evaluation experiments on a non-parallel many-to-many speaker identity conversion task
revealed that the proposed method obtained higher sound quality and speaker similarity than the VAEGAN-based method.

\small
\bibliographystyle{IEEEbib}
\bibliography{Kameoka2018arXiv08}

\end{document}